\documentclass[11pt,a4paper]{article}
\usepackage[hyperref]{emnlp2020}
\usepackage{times}
\usepackage{latexsym}

\usepackage{amsfonts}
\usepackage{amsmath}
\usepackage{url}
\usepackage{subfigure}
\usepackage{graphicx}
\usepackage{multirow}
\usepackage{multicol}
\usepackage{latexsym}
\usepackage{color}
\usepackage{xcolor}
\usepackage{stfloats}
\usepackage{microtype}

\aclfinalcopy 

\title{{A}n {I}nvestigation of {P}otential {F}unction {D}esigns for {N}eural {CRF}}

\author{Zechuan Hu$^{\diamond}$, Yong Jiang$^{\dagger *}$, Nguyen Bach$^{\dagger}$, Tao Wang$^{\dagger}$, Zhongqiang Huang$^{\dagger}$, \\ \textbf{ Fei Huang$^{\dagger}$, Kewei Tu}$^{\diamond}$\thanks{\hspace{1mm} Kewei Tu and Yong Jiang are the corresponding authors.} \\
 $^\diamond$School of Information Science and Technology, ShanghaiTech University \\
 Shanghai Engineering Research Center of Intelligent Vision and Imaging \\
 Shanghai Institute of Microsystem and Information Technology, Chinese Academy of Sciences \\
 University of Chinese Academy of Sciences \\
 $^\dagger$DAMO Academy, Alibaba Group \\
  {\tt \{huzch,tukw\}@shanghaitech.edu.cn} \\
  {\tt \{yongjiang.jy,nguyen.bach,leeo.wangt,f.huang\}@alibaba-inc.com} \\
 }

\date{}

\begin{document}
\maketitle
\begin{abstract}
The neural linear-chain CRF model is one of the most widely-used approach to sequence labeling. In this paper, we investigate a series of increasingly expressive potential functions for neural CRF models, which not only integrate the emission and transition functions, but also explicitly take the representations of the contextual words as input. Our extensive experiments show that the decomposed quadrilinear potential function based on the vector representations of two neighboring labels and two neighboring words consistently achieves the best performance. 
\end{abstract}

\section{Introduction}
Sequence labeling is the task of labeling each token of a sequence. It is an important task in natural language processing and has a lot of applications such as Part-of-Speech Tagging (POS) \cite{derose-1988-grammatical,toutanova-etal-2003-feature,xin-etal-2018-learning}, Named Entity Recognition (NER) \cite{ritter-etal-2011-named,akbik-etal-2019-pooled}, Chunking \cite{tjong-kim-sang-buchholz-2000-introduction,suzuki-etal-2006-training}.

The neural CRF model is one of the most widely-used approaches to sequence labeling and can achieve superior performance on many tasks \cite{collobert2011natural,chen-etal-2015-long,ling-etal-2015-finding,ma-hovy-2016-end,lample-etal-2016-neural}. 
It often employs an encoder such as a BiLSTM to compute the contextual vector representation of each word in the input sequence.
The potential function at each position of the input sequence in a neural CRF is typically decomposed into an emission function (of the current label and the vector representation of the current word) and a transition function (of the previous and current labels)  \cite{liu2018empower,yang-etal-2018-design}. 

In this paper, we design a series of increasingly expressive potential functions for neural CRF models. First, we compute the transition function from label embeddings \cite{ma-etal-2016-label,nam2016all,cui-zhang-2019-hierarchically} instead of label identities. Second, we use a single potential function over the current word and the previous and current labels, instead of decomposing it into the emission and transition functions, leading to more expressiveness. We also employ tensor decomposition in order to keep the potential function tractable. Thirdly, we take the representations of additional neighboring words as input to the potential function, instead of solely relying on the BiLSTM to capture contextual information. 

To empirically evaluate different approaches, we conduct experiments on four well-known sequence labeling tasks: NER, Chunking, coarse- and fine-grained POS tagging. We find that it is beneficial for the potential function to take representations of neighboring words as input, and a quadrilinear potential function with a decomposed tensor parameter leads to the best overall performance.

\begin{figure}[t]
    \centering
    \includegraphics[scale=0.52]{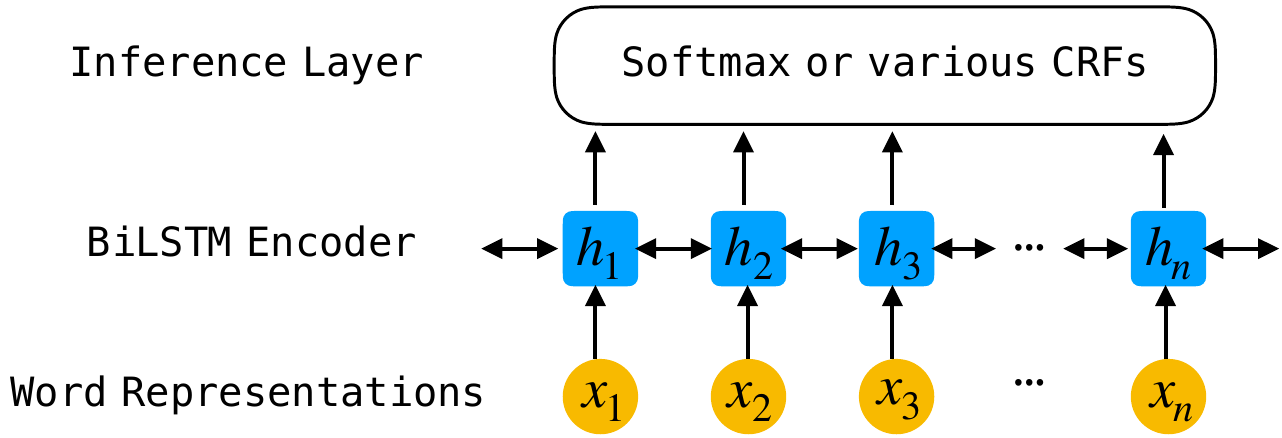}
    \caption{Neural architecture for sequence labeling }
    \label{fig:Network Architectures}
\end{figure}

Our work is related to \citet{reimers-gurevych-2017-reporting,yang-etal-2018-design}, which also compared different network architectures and configurations and conducted empirical analysis on different sequence labeling tasks. However, our focus is on the potential function design of neural CRF models, which has not been sufficiently studied before.

\section{Models}
Our overall neural network architecture for sequence labeling is shown in Figure \ref{fig:Network Architectures}. It contains three parts: a word representation layer, a bi-directional LSTM (BiLSTM) encoder, and an inference layer. The BiLSTM encoder produces a sequence of output vectors $\mathbf{h}_1, \mathbf{h}_2, ... \mathbf{h}_M \in \mathbb{R}^{D_h}$, which are utilized by the inference layer to predict the label sequence. The inference layer typically defines a potential function $s(\mathbf{x}, \mathbf{y}, i)$ for each position $i$ of the input sequence $\mathbf{x}$ and label sequence $\mathbf{y}$ and computes the conditional probability of the label sequence given the input sequence as follows:
\begin{align*}
   P(\mathbf{y}|\mathbf{x}) = \frac{\exp(\sum_{i=1}^{M} s(\mathbf{x}, \mathbf{y}, i))}{\sum_{\mathbf{y}'} \exp(\sum_{i=1}^{M} s(\mathbf{x}, {\mathbf{y}}', i))}
\end{align*}
where $M$ is the length of the sequence.

\begin{figure*}
    \centering
    \includegraphics[scale=0.66]{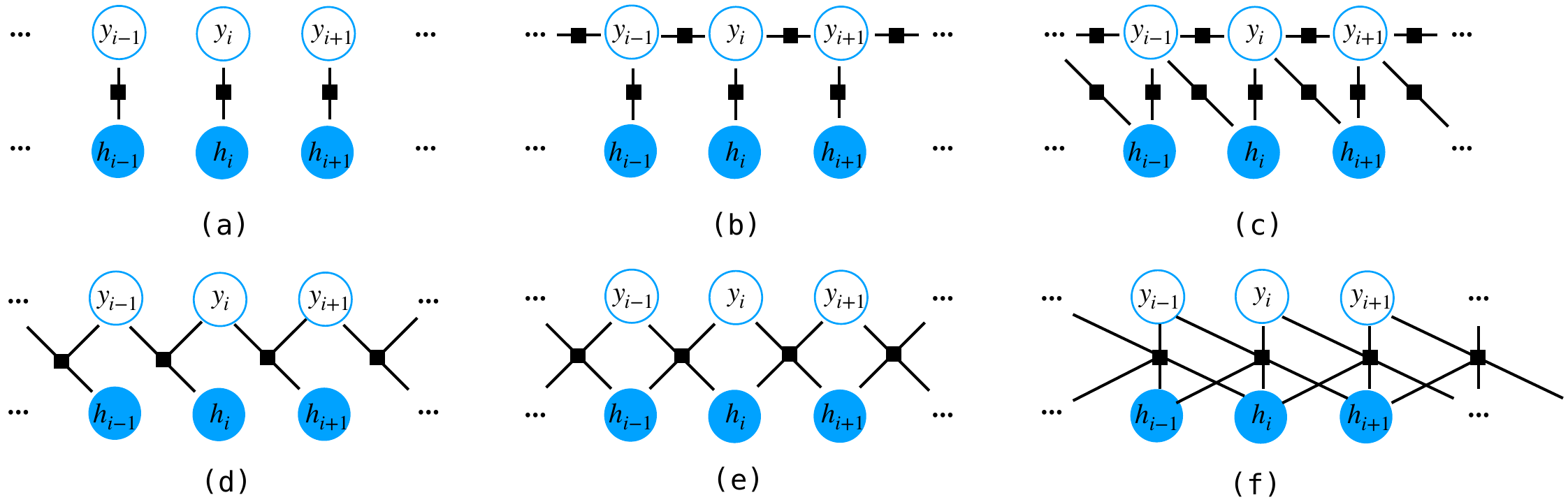}
    \caption{Factor graphs of different models. The solid circles and hollow circles indicate random variables of word encodings and labels respectively. The black squares represent factors.}
    \label{fig:Factor Graphs}
\end{figure*}

The simplest inference layer assumes independence between labels. It applies a linear transformation to $\mathbf{h}_i$ followed by a \textbf{Softmax} function to predict the distribution of label $y_i$ at each position $i$ (Figure \ref{fig:Factor Graphs}(a)). In many scenarios, however, it makes sense to model dependency between neighboring labels, which leads to linear-chain CRF models.

\paragraph{Vanilla CRF} In most previous work of neural CRFs, the potential function is decomposed to an emission function and a transition function (Figure \ref{fig:Factor Graphs}(b)), and the transition function is represented by a table $\phi$ maintaining the transition scores between labels. 
\begin{align*}
    s(\mathbf{x}, \mathbf{y}, i)&= \mathbf{v}_{y_{i-1}}^{T} \phi \mathbf{v}_{y_{i}} + \mathbf{h}_i^{T} \mathbf{W}_h \mathbf{v}_{y_i}
\end{align*}
where $\mathbf{v}_{y_i}$ is a one-hot vector for label $y_{i}$ and $\mathbf{W}_{h} \in \mathbb{R}^{D_h \times D_t}$ is a weight matrix.

\paragraph{TwoBilinear} Instead of one-hot vectors, we may use dense vectors to represent labels, which has the benefit of encoding similarities between labels. Accordingly, the emission and transition functions are modeled by two bilinear functions.
\begin{align*}
    s(\mathbf{x}, \mathbf{y}, i) = \mathbf{t}_{y_{i-1}}^{T} \mathbf{W}_{t} \mathbf{t}_{y_i} + \mathbf{h}_i^{T} \mathbf{W}_{h} \mathbf{t}_{y_i}
\end{align*}
where $\mathbf{W}_{t} \in \mathbb{R}^{D_t \times D_t}$ is a weight matrix, and $\mathbf{t}_{y_i} \in \mathbb{R}^{D_t}$ is the embedding of label $y_i$. The factor graph remains the same as vanilla CRF (Figure \ref{fig:Factor Graphs}(b)).

\paragraph{ThreeBilinear} Figure \ref{fig:Factor Graphs}(c) depicts the structure of ThreeBilinear. Compared with TwoBilinear, ThreeBilinear has an extra emission function between the current word representation and previous label.
\begin{align*}
    s(\mathbf{x}, \mathbf{y}, i) =  & \mathbf{t}_{y_{i-1}}^{T} \mathbf{W}_{t} \mathbf{t}_{y_i}  + \mathbf{h}_i^{T} \mathbf{W}_{h_1} \mathbf{t}_{y_i} \\
    &  +  \mathbf{h}_i^{T} \mathbf{W}_{h_2} \mathbf{t}_{y_{i-1}}
\end{align*}

\paragraph{Trilinear} Instead of three bilinear functions, we may use a trilinear function to model the correlation between $\mathbf{h}_i$, $\mathbf{t}_{y_i}$ and $\mathbf{t}_{y_{i-1}}$. It has strictly more representational power than the sum of three bilinear functions.
\begin{align*}
    s(\mathbf{x}, \mathbf{y}, i) = \mathbf{h}_i^{T} \mathbf{U} \mathbf{t}_{y_{i-1}} \mathbf{t}_{y_i}
\end{align*}
where $\mathbf{U} \in \mathbb{R}^{D_h \times D_t \times D_t}$ is an order-3 weight tensor. Figure \ref{fig:Factor Graphs}(d) presents the structure of Trilinear.  

\paragraph{D-Trilinear} Despite the increased representational power of Trilinear, its space and time complexity becomes cubic.
To reduce the computational complexity without too much compromise of the representational power, we assume that $\mathbf{U}$ has rank $D_r$ and can be decomposed into the product of three matrices $\mathbf{U}_{t_1}$, $\mathbf{U}_{t_2} \in \mathbb{R}^{D_t \times D_r}$ and $\mathbf{U_h} \in \mathbb{R}^{D_h \times D_r}$. Then the trilinear function can be rewritten as,
\begin{align*}
    & s(\mathbf{x}, \mathbf{y}, i) = \sum_{j=1}^{D_r} (\mathbf{g}_1 \circ \mathbf{g}_2 \circ \mathbf{g}_3)_j \\
    & \mathbf{g}_1 = \mathbf{t}_{y_{i-1}}^T \mathbf{U}_{t_1};\quad\mathbf{g}_2 = \mathbf{t}_{y_{i}}^T \mathbf{U}_{t_2};\quad\mathbf{g}_3 = \mathbf{h}_i^T \mathbf{U}_{h}
\end{align*}
where $\circ$ denotes element-wise product. We call the resulting model D-Trilinear. The factor graph of D-Trilinear is the same as Trilinear (Figure \ref{fig:Factor Graphs}(d)).

\paragraph{D-Quadrilinear} We may take the representation of the previous word as an additional input and use a quadrilinear function in the potential function.
\begin{align*}
    s(\mathbf{x}, \mathbf{y}, i) = \mathbf{h}_{i-1}^{T} \mathbf{h}_i^{T} \mathbf{U} \mathbf{t}_{y_{i-1}} \mathbf{t}_{y_i}
\end{align*}
where $\mathbf{U}$ is an order-4 weight tensor. 
However, the computational complexity of this function becomes quartic. Hence we again decompose the tensor into the product of four matrices and rewrite the potential function as follows.
\begin{align*}
    & s(\mathbf{x}, \mathbf{y}, i) = \sum_{j=1}^{D_r} (\mathbf{g}_1 \circ \mathbf{g}_2 \circ \mathbf{g}_3 \circ \mathbf{g}_4)_j \\
    & \mathbf{g}_1 = \mathbf{t}_{y_{i-1}}^T \mathbf{U}_{t_1};\quad\mathbf{g}_2 = \mathbf{t}_{y_i}^T \mathbf{U}_{t_2} \\
    & \mathbf{g}_3 = \mathbf{h}_{i-1}^T \mathbf{U}_{h_1};\quad\mathbf{g}_4 = \mathbf{h}_{i}^T \mathbf{U}_{h_2}
\end{align*}
We call the resulting model D-Quadrilinear and its factor graph is shown in Figure \ref{fig:Factor Graphs}(e).

\paragraph{D-Pentalinear} Following the same idea, we extend D-quadrilinear to D-Pentalinear by taking the representation of the next word as an additional input.
Figure \ref{fig:Factor Graphs}(f) shows the structure of D-Pentalinear. 
\begin{align*}
    & s(\mathbf{x}, \mathbf{y}, i) = \sum_{j=1}^{D_r} (\mathbf{g}_1 \circ \mathbf{g}_2 \circ \mathbf{g}_3 \circ \mathbf{g}_4 \circ \mathbf{g}_5)_j \\
    & \mathbf{g}_1 = \mathbf{t}_{y_{i-1}}^T \mathbf{U}_{t_1};\quad\mathbf{g}_2 = \mathbf{t}_{y_i}^T \mathbf{U}_{t_2} \\
    & \mathbf{g}_3 = \mathbf{h}_{i-1}^T \mathbf{U}_{h_1};\quad\mathbf{g}_4 =\mathbf{h}_{i}^T\mathbf{U}_{h_2};\quad \mathbf{g}_5 = \mathbf{h}_{i+1}^T \mathbf{U}_{h_3}
\end{align*}

\begin{table*}
\setlength\tabcolsep{3.8pt}
\renewcommand\arraystretch{1.46}
\scriptsize
\centering
\begin{tabular}{l|l|ccccc||cccc}
\multicolumn{2}{c|}{}& \multicolumn{5}{c||}{\bf \textsc{NER}} & \multicolumn{4}{c}{\bf \textsc{Chunking}}\\
\multicolumn{2}{c|}{}& \textbf{English} & \textbf{German} &  \textbf{Dutch}& \textbf{Spanish} & \bf Avg. &  \textbf{English}&\textbf{German} &  \textbf{Vietnamese}& \bf Avg.\\ 

\hline \hline
\multirow{7}{*}{\rotatebox{90}{\bf \textsc{BERT Embedding}}}	
& \bf \textsc{Softmax} &	90.42$\pm$0.16&	81.91$\pm$0.15&	89.02$\pm$0.31&	85.86$\pm$0.34&	86.80$\pm$0.24&	90.72$\pm$0.11&	93.48$\pm$0.07&	74.13$\pm$0.27&	86.11$\pm$0.15\\
& \bf \textsc{Vanilla CRF} &	91.33$\pm$0.18&	83.56$\pm$0.18&	90.03$\pm$0.18&	87.32$\pm$0.38&	88.06$\pm$0.23&	91.05$\pm$0.12&	93.65$\pm$0.08&	76.07$\pm$0.08&	86.92$\pm$0.09\\	\cline{2-11}
										
& \bf \textsc{TwoBilinear} &	91.23$\pm$0.07&	83.21$\pm$0.35&	90.02$\pm$0.26&	87.40$\pm$0.24&	87.96$\pm$0.23&	91.16$\pm$0.04&	93.60$\pm$0.10&	76.10$\pm$0.23&	86.95$\pm$0.13\\
& \bf \textsc{ThreeBilinear} &	91.19$\pm$0.24&	83.35$\pm$0.19&	90.06$\pm$0.45&	87.38$\pm$0.18&	87.99$\pm$0.26&	91.13$\pm$0.14&	93.52$\pm$0.14&	75.98$\pm$0.23&	86.87$\pm$0.17\\
& \bf \textsc{Trilinear} &	91.24$\pm$0.11&	83.11$\pm$0.27&	90.53$\pm$0.41&	87.38$\pm$0.26&	88.07$\pm$0.26&	91.11$\pm$0.04&	93.68$\pm$0.09&	75.64$\pm$0.25&	86.81$\pm$0.13\\	\cline{2-11}
										
& \bf \textsc{D-Trilinear} &	91.28$\pm$0.16&	83.25$\pm$0.36&	90.52$\pm$0.25&	87.68$\pm$0.13&	88.18$\pm$0.22&	91.32$\pm$0.08&	93.79$\pm$0.10&	76.18$\pm$0.13&	87.10$\pm$0.10\\
& \bf \textsc{D-Quadrilinear} &	91.46$\pm$0.07&83.61$\pm$0.22&	\textbf{90.76$\pm$0.13}&	\textbf{87.71$\pm$0.29}&	\textbf{88.38$\pm$0.18}&	\textbf{91.51$\pm$0.11}&	94.08$\pm$0.08&	\textbf{76.29$\pm$0.36}&	\textbf{87.29$\pm$0.18}\\
& \bf \textsc{D-Pentalinear} &\textbf{91.47$\pm$0.20}&\textbf{83.63$\pm$0.26}&	90.50$\pm$0.27&	87.69$\pm$0.20&	88.33$\pm$0.23&	91.45$\pm$0.08&	\textbf{94.23$\pm$0.06}&	76.01$\pm$0.20&	87.23$\pm$0.11\\
\end{tabular}
\caption{\label{table:NER & Chunking Result}
Results on NER and Chunking with BERT embeddings. 
}
\end{table*}

\section{Experiments}
We compare neural Softmax and the seven variants of neural CRFs on four sequence labeling tasks: NER, Chunking, coarse- and fine-grained POS tagging. 
For NER, we use the datasets from CoNLL 2002 and 2003 shared tasks \cite{tjong-kim-sang-2002-introduction,tjong-kim-sang-de-meulder-2003-introduction}.
For Chunking, we use the English and German datasets of the CoNLL 2003 shared task \cite{tjong-kim-sang-de-meulder-2003-introduction} and the Vietnamese dataset \cite{pham-etal-2017-nnvlp}. For the two POS tagging tasks, we select 8 languages from Universal Dependencies (UD \textbf{v2.4}) treebanks \cite{11234/1-2988}.

We conduct our experiments with pretrained word embeddings, character embeddings, and BERT embeddings \cite{devlin-etal-2019-bert}. For NER and Chunking, we use the BIOES scheme for its better performance than the BIO scheme \cite{ratinov-roth-2009-design,dai2015enhancing,yang-etal-2018-design}. We use F1-score as the evaluation metric for both NER and Chunking. We run each model for 5 times with different random seeds for each experiment and report the average score and the standard derivation. More details can be found in supplementary material.

\begin{table*}
\setlength\tabcolsep{2.3pt}
\renewcommand\arraystretch{1.36}
\scriptsize
\centering
\begin{tabular}{l|cc||cc||ccc||ccc}
    & \multicolumn{2}{c||}{\bf \textsc{NER}}&\multicolumn{2}{c||}{\bf\textsc{Chunking}}&\multicolumn{3}{c||}{\bf\textsc{Fine-grained POS}}&\multicolumn{3}{c}{\bf\textsc{Coarse-grained POS}}\\
    &\bf  \textsc{word} &\bf \textsc{char}&\bf \textsc{word}&\bf \textsc{char} &\bf  \textsc{word} &\bf \textsc{char}&\bf \textsc{bert}&\bf  \textsc{word} &\bf \textsc{char}&\bf \textsc{bert} \\
        \hline \hline
    {\bf \textsc{BiLSTM-LAN}}&77.70$\pm$0.39&82.42$\pm$0.55&85.59$\pm$0.12&86.12$\pm$0.12&94.45$\pm$0.14&95.41$\pm$0.13&---&94.75$\pm$0.10&95.68$\pm$0.08&---\\
    \hline
    {\bf \textsc{Softmax}}&78.22$\pm$0.32&82.14$\pm$0.26&84.99$\pm$0.14&85.49$\pm$0.07&94.91$\pm$0.08&95.72$\pm$0.07&95.83$\pm$0.07
    &94.47$\pm$0.09&95.58$\pm$0.08&96.18$\pm$0.08\\
    {\bf \textsc{Vanilla CRF}} &79.46$\pm$0.57&83.59$\pm$0.66&85.86$\pm$0.11&86.39$\pm$0.08&94.89$\pm$0.08&95.70$\pm$0.11&95.81$\pm$0.09&94.53$\pm$0.10&95.60$\pm$0.10&96.23$\pm$0.09\\
    \hline 
    {\bf \textsc{TwoBilinear}} &79.16$\pm$0.42&83.36$\pm$0.42&85.57$\pm$0.19&85.94$\pm$0.15&94.81$\pm$0.11&95.64$\pm$0.10&95.79$\pm$0.09&94.48$\pm$0.08&95.58$\pm$0.11&96.18$\pm$0.09\\
    {\bf \textsc{ThreeBilinear}} &78.66$\pm$0.94&83.53$\pm$0.28&85.51$\pm$0.23&85.95$\pm$0.21&94.87$\pm$0.09&95.66$\pm$0.09&95.74$\pm$0.11&94.49$\pm$0.09&95.54$\pm$0.09&96.14$\pm$0.08\\
    \multirow{1}{*}{\bf \textsc{Trilinear}} &79.24$\pm$0.35&83.50$\pm$0.38&85.57$\pm$0.28&86.08$\pm$0.31&94.94$\pm$0.13&95.71$\pm$0.11&95.67$\pm$0.11&94.61$\pm$0.11&95.63$\pm$0.12&96.17$\pm$0.14\\
    \hline
    {\bf \textsc{D-Trilinear}} &79.41$\pm$0.24&83.75$\pm$0.39&85.83$\pm$0.13&86.42$\pm$0.14&95.07$\pm$0.10&95.75$\pm$0.08&95.74$\pm$0.11&94.70$\pm$0.11&95.69$\pm$0.08&96.25$\pm$0.08\\
    {\bf \textsc{D-Quadrilinear}} &\textbf{80.09$\pm$0.35}&\textbf{84.20$\pm$0.39}&\textbf{86.58$\pm$0.14}&87.07$\pm$0.10&\textbf{95.19$\pm$0.08}&\textbf{95.88$\pm$0.08}&\textbf{95.90$\pm$0.09}&\textbf{94.91$\pm$0.10}&\textbf{95.82$\pm$0.10}&\textbf{96.32$\pm$0.07}\\
    {\bf \textsc{D-Pentalinear}} &79.52$\pm$0.28&84.01$\pm$0.42&86.53$\pm$0.15&\textbf{87.11$\pm$0.20}&95.07$\pm$0.19&95.82$\pm$0.11&95.85$\pm$0.08&94.80$\pm$0.15&95.79$\pm$0.13&96.31$\pm$0.11\\
\end{tabular}
\caption{Results averaged over all the languages for each task. We also show the results of BiLSTM-LAN \cite{cui-zhang-2019-hierarchically}, one of the current state-of-the-art sequence labeling approaches, for reference.\label{table:average of char and word} We do not report the results of BiLSTM-LAN with BERT embedding because BERT is not available in the BiLSTM-LAN code.}
\end{table*}

\subsection{Results}\label{sec:Results}
We show the detailed results on NER and Chunking with BERT embeddings in Table \ref{table:NER & Chunking Result} and the averaged results on all the tasks in Table \ref{table:average of char and word} (the complete results can be found in the supplementary materials). 
We make the following observations.
\textbf{Firstly}, D-Quadrilinear has the best overall performance in all the tasks. Its advantage over D-Trilinear is somewhat surprising because the BiLSTM output $\mathbf{h}_i$ in D-Trilinear already contains information of both the current word and the previous word. We speculate that: 1) information of the previous word is useful in evaluating the local potential in sequence labeling (as shown by traditional feature-based approaches); and 2) information of the previous word is obfuscated in $\mathbf{h}_i$ and hence directly inputting $\mathbf{h}_{i-1}$ into the potential function helps.  \textbf{Secondly}, D-Quadrilinear greatly outperforms BiLSTM-LAN \cite{cui-zhang-2019-hierarchically}, one of the state-of-the-art sequence labeling approaches which employs a hierarchically-refined label attention network. \textbf{Thirdly}, D-Trilinear clearly outperforms both ThreeBilinear and Trilinear. This suggests that tensor decomposition could be a viable way to both regularize multilinear potential functions and reduce their computational complexity.

\begin{table}[tph]
\setlength\tabcolsep{7pt}
\renewcommand\arraystretch{1.36}
\scriptsize
\centering
\begin{tabular}{ll|c|c}
\multicolumn{2}{c|}{} & \multicolumn{1}{c|}{\bf \textsc{NER}} & \multicolumn{1}{c}{\bf \textsc{Chunking}}\\
\hline \hline

\multirow{3}{*}{\bf \textsc{Layers=2}} 
&{\bf \textsc{Vanilla CRF}}&79.86$\pm$0.47&85.84$\pm$0.19\\
&{\bf \textsc{D-Trilinear}} &80.21$\pm$0.34&85.86$\pm$0.19 \\
&{\bf \textsc{D-Quadrilinear}} &\textbf{80.36$\pm$0.34}&\textbf{86.32$\pm$0.14} \\
\hline \hline
\multirow{3}{*}{\bf \textsc{Layers=3}}
&{\bf \textsc{Vanilla CRF}}
&78.72$\pm$0.66&85.73$\pm$0.15\\
&{\bf \textsc{D-Trilinear}} &79.84$\pm$0.62&85.65$\pm$0.20\\
&{\bf \textsc{D-Quadrilinear}} &\textbf{79.97$\pm$0.31}&\textbf{85.88$\pm$0.15}\\
\end{tabular}
\caption{Average results with more BiLSTM layers.\label{table:Multi-Layers}}
\end{table}

\begin{figure}
    \centering
    \includegraphics[scale=0.6]{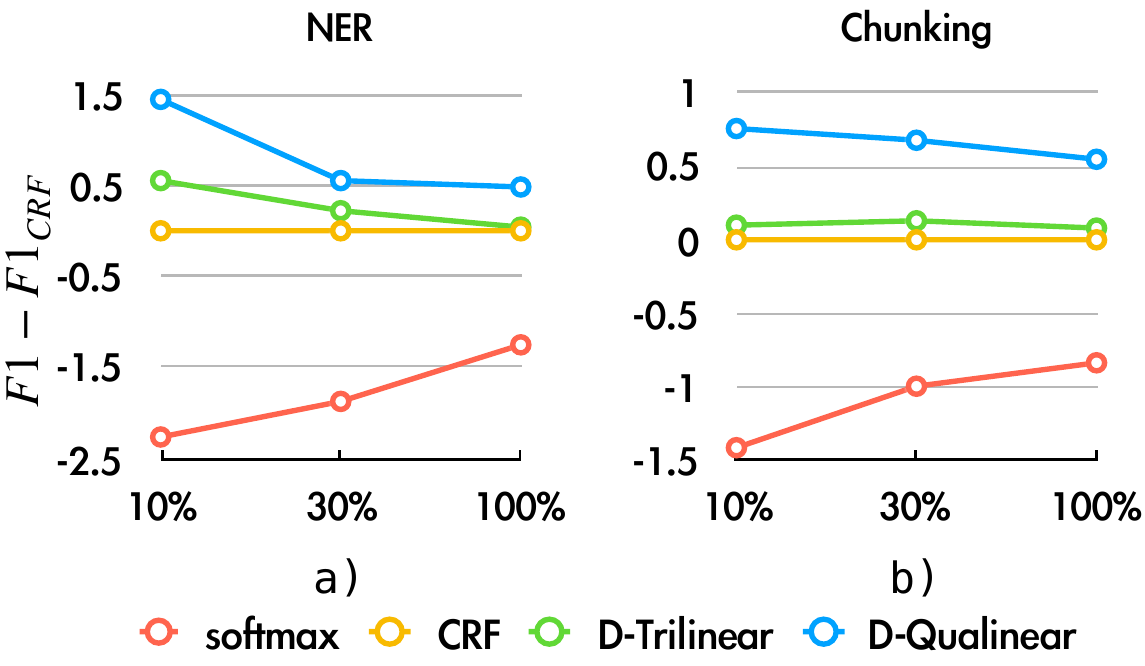}
    
    \caption{The average differences in F1-scores compared with Vanilla CRF with different training data sizes. }
    \label{fig:Small Data}
\end{figure}

\subsection{Analysis}

\paragraph{Small training data } We train four of our models on randomly selected 10\% or 30\% of the training data on the NER and Chunking tasks. We run each experiment for 5 times. Figure \ref{fig:Small Data} shows the average difference in F1-scores between each model and Vanilla CRF.  It can be seen that with small data, the advantages of D-Trilinear and D-Quadrilinear over Vanilla CRF and Softmax become even larger.

\paragraph{Multi-layers LSTM} As discussed in section \ref{sec:Results}, D-Quadrilinear outperforms D-Trilinear probably because $\mathbf{h}_i$, the BiLSTM output at position $i$, does not contain sufficient information of the previous word. Here we study whether increasing the number of BiLSTM layers would inject more information into $\mathbf{h}_i$ and hence reduce the performance gap between the two models. Table \ref{table:Multi-Layers} shows the results on the NER and Chunking tasks with word embedding. D-Quadrilinear still outperforms D-Trilinear, but by comparing Table \ref{table:Multi-Layers} with Table \ref{table:average of char and word}, we see that their difference indeed becomes smaller with more BiLSTM layers.
Another observation is that more BiLSTM layers often lead to lower scores. This is consistent with previous findings \cite{cui-zhang-2019-hierarchically} and is probably caused by overfitting.

\paragraph{Speed} We test the training and inference speed of our models. Our decomposed multilinear approaches are only a few percent slower than Vanilla CRF during training and as fast as Vanilla CRF during inference, which suggests their practical usefulness. The details can be found in the supplementary material.

\section{Conclusion}
In this paper, we investigate several potential functions for neural CRF models. The proposed potential functions not only integrate the emission and transition functions, but also take into consideration representations of additional neighboring words. Our experiments show that D-Quadrilinear achieves the best overall performance. Our proposed approaches are simple and effective and could facilitate future research in neural sequence labeling.

\section*{Acknowledgement}
This work was supported by Alibaba Group through Alibaba Innovative Research Program. This work was also supported by the National Natural Science Foundation of China (61976139). 
\bibliographystyle{acl_natbib}
\bibliography{Quadrilinear,anthology}

\appendix
\label{sec:appendix}
\section{Appendices}

\subsection{Dataset Statistics}
\label{sec:Statistics}

The statistics of the datasets used in our experiments are listed in table \ref{table:statistics}.
\begin{table}[ht]
\small
\centering
\begin{tabular}{llllll}
\textbf{Task} & \textbf{D} & \textbf{\#train} & \textbf{\#dev} & \textbf{\#test} & \textbf{\#label} \\
\hline \hline
\multirow{4}{*}{\textbf{NER}} & en & 14040 & 3250 & 3453 & 17\\
             & de & 12152 & 2867 & 3005 & 17 \\
             & nl & 15796 & 2895 & 5196 & 17 \\
             & sp & 8319 & 1914 & 1517 & 17\\
\hline
\multirow{3}{*}{\textbf{Chunking}} & en & 14040 & 3250 & 3453 & 38\\
             & de & 12152 & 2867 & 3005 & 13 \\
             & vi & 6284 & 786 & 785 & 37\\
\hline
\multirow{8}{*}{\textbf{POS}} & en & 12543 & 2002 & 2077 &50/17\\
             & de & 13814 & 799 & 977 &52/17\\
             & it & 13121 & 564 & 482 &39/17\\
             & id & 4477 & 559 & 557 &81/16\\
             & nl & 12269 & 718 & 596 &194/16\\
             & hi & 13304 & 1659 & 1684 &31/16\\
             & zh & 3997 & 500 & 500 &42/15\\
             & ja & 7125 & 511 & 550 &37/16\\
\end{tabular}
\caption{\label{table:statistics}
The statistics of different datasets for corresponding tasks. D: Datasets. The statistics of Coarse-grained POS is the same as Fine-grained POS except that the number of labels are not the same. The left of `/` indicates the number of labels of Fine-grained POS and the right of  `/` indicates the number of labels of Coarse-grained POS.
}
\end{table}

\subsection{Word representations} 
We have three different versions of word representations:
\begin{itemize}
    \item \textbf{Word Embedding.} We use pretrained word embeddings such as GloVe \cite{pennington2014glove} and FastText \cite{Grave2018LearningWV}.
    \item \textbf{Word Embedding and Character Embedding.} We use the same character LSTMs as in \citet{lample2016neural} and set the hidden size of the LSTM to 50. The final word representation is the concatenation of the output of the character LSTM and the pretrained word embedding.
    \item \textbf{BERT Embedding.} We use the respective BERT embedding from \cite{Devlin2019BERTPO} for each language. If there is no pretrained BERT embedding for a language, we then use the multilingual BERT (M-BERT) instead. The word representation is from the last four layers of the BERT embedding.
\end{itemize}
We fine-tune the word embeddings and character embeddings during the training process. We don't fine-tune the BERT embeddings.

\begin{table}
\normalsize
\centering
\begin{tabular}{ll}

\textbf{Hyperparameters} & \textbf{Setting}\\ 
\hline \hline
LSTM Hidden Size & 512 \\
Learning Rate & 0.1 \\
Char Embeddng Size & 25 \\
Char Hidden Size & 50 \\
Dropout Rate & 0.5 \\ 
L2 Regularization & 1e-8\\
Batch Size & 32 \\
Maximal Epochs & 300 \\
Patience & 10
\end{tabular}
\caption{\label{table:parameter settings}
Other hyperparameters
}
\end{table}

\subsection{Hyperparameters setting}
We tune the following hyperparameters in our experiments.

\paragraph{LSTM hidden size} We test Softmax, Vanilla CRF, D-Trilinear and D-Quadlinear with LSTM hidden sizes of \{200, 512\} on the English and German datasets of each task and find that there is no significant difference between 200 and 512. Hence, we fix the LSTM hidden size to 512.

\paragraph{Learning Rate} We tune it in the range of \{0.03, 0.1, 0.3\} on Softmax, Vanilla CRF, D-Trilinear and D-Quadlinear on the English and German datasets of each task. We find that the performance is always better when the learning rate is $0.1$. So we fix the learning rate to 0.1.

\paragraph{Tag Embedding Dimension $D_t$} We use tag embeddings in all the models except Softmax and Vanilla CRF. We search for the best dimension in \{20, 50, 100, 200\}.

\paragraph{Rank $D_r$} In D-Trilinear, D-Quadlinear, and D-Pentalinear, $D_r$ is a hyperparameter that controls the representational power of the multilinear functions. We select its value from \{64, 128, 256, 384, 600\}.

Other hyperparameter settings are list in table \ref{table:parameter settings}.

\makeatletter 
  \newcommand\tabcaption{\def\@captype{table}\caption} 
\makeatother
\begin{figure*}
\minipage{.54\textwidth}
  \setlength\tabcolsep{1.75pt}
\renewcommand\arraystretch{1.562}
\scriptsize
\centering
\begin{tabular}{l|ccccc}
    &\textbf{English} & \textbf{German} &  \textbf{Dutch}& \textbf{Spanish}& \textbf{Avg.}\\
    \hline \hline
    \multirow{1}{*}{\bf \textsc{TwoBilinear}} &90.11$\pm$0.25&73.69$\pm$0.30&69.31$\pm$0.67&83.53$\pm$0.44&79.16$\pm$0.42\\
    \multirow{1}{*}{\bf \textsc{ThreeBilinear}} &90.12$\pm$0.22&73.20$\pm$0.85&67.50$\pm$2.35&	83.82$\pm$0.32&78.66$\pm$0.94\\
    \multirow{1}{*}{\bf \textsc{Trilinear}} &90.19$\pm$0.19&73.39$\pm$0.40&69.69$\pm$0.60&	83.70$\pm$0.19&79.24$\pm$0.35\\
    \multirow{1}{*}{\bf \textsc{D-Trilinear}} &90.43$\pm$0.23&73.57$\pm$0.17&69.50$\pm$0.32&84.15$\pm$0.23&79.41$\pm$0.24\\
    \multirow{1}{*}{\bf \textsc{1Word+2label}}
    &89.91$\pm$0.04&74.37$\pm$0.12&68.96$\pm$0.83&83.68$\pm$0.22&79.23$\pm$0.30\\
    \hline 
    \multirow{1}{*}{\bf \textsc{D-Quadlinear}} 
    &\textbf{90.44$\pm$0.07}&\textbf{75.05$\pm$0.35}&\textbf{70.49$\pm$0.68}&\textbf{84.41$\pm$0.29}&\textbf{80.10$\pm$0.35}\\
    \multirow{1}{*}{\bf \textsc{2Word+2label}}
    &90.27$\pm$0.09&74.19$\pm$0.32&70.34$\pm$0.06&83.81$\pm$0.37&79.65$\pm$0.21\\
\end{tabular}
\tabcaption{Comparison with concatenation-based potential functions (\textsc{1Word+2Label} and \textsc{2Word+2Label})\label{table:Concat}}
\endminipage\hfill
\minipage{.44\textwidth}
  \includegraphics[width=1\linewidth]{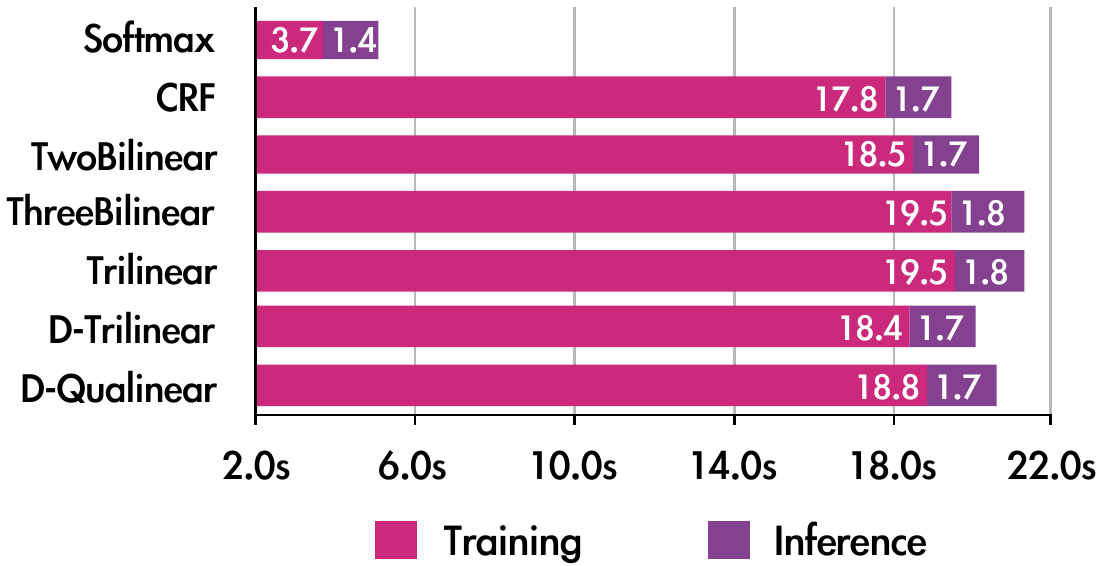}
  \caption{Training time and inference time averaged over 10 epochs.}    
  \label{fig:Speed Test}
\endminipage
\end{figure*}

\begin{table}
\setlength\tabcolsep{1.75pt}
\renewcommand\arraystretch{1.16}
\scriptsize
\centering
\begin{tabular}{l|ccccc}
\multicolumn{1}{c|}{}& \textbf{English} & \textbf{German} &  \textbf{Dutch}& \textbf{Spanish} & \bf Avg. \\
\hline \hline

{\bf \textsc{Vanilla CRF}} & 88.33 & 78.59 & 64.88 &  81.2 & 78.25\\
{\bf \textsc{D-Quadrilinear}} &\textbf{89.49}&\textbf{79.93}&\textbf{67.23}&\textbf{81.6}&\textbf{79.56}\\
\end{tabular}
\caption{Average results with transformer encoder.\label{table:transformer_result}}
\end{table}

\subsection{Additional Analysis}



\paragraph{Multilinear vs. Concatenation} Our best-performing models are based on multilinear functions with decomposed parameter tensors. An alternative to multilinear functions is to apply an MLP with nonlinear activations to the concatenated input vectors.
We run the comparison on the NER task with word embeddings and tune the tag embedding size from \{20, 50, 100, 200\} and the hidden size of the MLP from \{64, 128, 256, 384\}. As shown in table \ref{table:Concat},  the two concatenation-based models underperform their decomposed multilinear counterparts, but they do outperform TwoBilinear and ThreeBilinear.

\paragraph{Transformer vs. BiLSTM}
As we discussed in section \ref{sec:Results}, information of the previous word may be obfuscated in $\mathbf{h}_i$. Transformer-like encoders which can model long-range context may alleviate the obfuscation. We use a 6-layers transformer encoder and run the comparison on vanilla CRF and D-Quadrilinear on NER tasks with word embeddings. As shown in table \ref{table:transformer_result}, with the transformer encoder, D-Quadrilinear outperforms the vanilla CRF by 1.31\%. In comparison, with the BiLSTM encoder, D-Quadrilinear outperforms the vanilla CRF by 0.63\%. So the advantage of our approach against the vanilla CRF becomes even larger when using the transformer encoder.

\paragraph{Speed} We use a Nvidia Titan V GPU to test the training and inference speed of the 8 models on the NER English dataset. Figure \ref{fig:Speed Test} shows the training and inference time averaged over 10 epochs. Softmax is much faster than all the other approaches because it does not need to run Forward-Backward and Viberbi and can parallelize the predictions at all the positions of a sequence. Our decomposed multilinear approaches are not significantly slower than Vanilla CRF but generally have better performance, which suggests their practical usefulness.

\subsection{Complete Experimental Results}
Table \ref{table:details of NER & Chunking Result}, \ref{table:details of Coarse-grained POS Result}, and \ref{table:details of Fine-grained POS Result} show the detailed results on the NER, Chunking and two POS tasks. 

In addition, we show results of BiLSTM-LAN \cite{cui-zhang-2019-hierarchically}, which is one of the state-of-the-art sequence labeling approaches. We run the released code of BiLSTM-LAN\footnote{\url{https://github.com/Nealcly/BiLSTM-LAN}} on NER, Chunking and the two POS tagging tasks. We tune BiLSTM-LAN hyperparameters with the word-level hidden size of \{100, 200, 400\}, LSTM layer number of \{1, 2, 3, 4\}, learning rate of \{0.003, 0.01, 0.03\}, and decay rate of \{0.03, 0.035, 0.04\}. All the other hyperparameters follow their default settings.
We do not report results of BiLSTM-LAN with BERT embedding because BERT is not available in the BiLSTM-LAN code.


\begin{table*}[ht]
\setlength\tabcolsep{3.8pt}
\renewcommand\arraystretch{1.46}
\scriptsize
\centering
\begin{tabular}{l|l|ccccc||cccc}
\multicolumn{2}{c|}{}& \multicolumn{5}{c||}{\bf \textsc{NER}} & \multicolumn{4}{c}{\bf \textsc{Chunking}}\\
\multicolumn{2}{c|}{}& \textbf{English} & \textbf{German} &  \textbf{Dutch}& \textbf{Spanish} & \bf Avg. &  \textbf{English}&\textbf{German} &  \textbf{Vietnamese}& \bf Avg.\\ 
\hline \hline
\multirow{8}{*}{\rotatebox{90}{\bf \textsc{Word Embedding}}}& \bf \textsc{BiLSTM-LAN}&89.46$\pm$0.24&72.48$\pm$0.35&66.02$\pm$0.70&82.83$\pm$0.26&77.70$\pm$0.39&91.46$\pm$0.11&93.16$\pm$0.10&72.16$\pm$0.14&85.59$\pm$0.12\\\cline{2-11}
& \bf \textsc{Softmax} &	89.69$\pm$0.13&	72.59$\pm$0.30&	68.01$\pm$0.37&	82.60$\pm$0.45&	78.22$\pm$0.32&	90.77$\pm$0.17&	93.04$\pm$0.09&	71.17$\pm$0.17&	84.99$\pm$0.14\\
& \bf \textsc{Vanilla CRF} &	90.33$\pm$0.36&	73.96$\pm$0.26&	69.72$\pm$1.00&	83.81$\pm$0.67&	79.46$\pm$0.57&	91.19$\pm$0.12&	93.15$\pm$0.09&	73.23$\pm$0.12&	85.86$\pm$0.11\\	\cline{2-11}
										
& \bf \textsc{TwoBilinear} &	90.11$\pm$0.25&	73.69$\pm$0.30&	69.31$\pm$0.67&	83.53$\pm$0.44&	79.16$\pm$0.42&	91.45$\pm$0.09&	92.98$\pm$0.09&	72.28$\pm$0.39&	85.57$\pm$0.19\\
										
& \bf \textsc{ThreeBilinear} &	90.12$\pm$0.22&	73.20$\pm$0.85&	67.50$\pm$2.35&	83.82$\pm$0.32&	78.66$\pm$0.94&	91.35$\pm$0.24&	92.98$\pm$0.07&	72.21$\pm$0.38&	85.51$\pm$0.23\\
										
& \bf \textsc{Trilinear} &	90.19$\pm$0.19&	73.39$\pm$0.40&	69.69$\pm$0.60&	83.70$\pm$0.19&	79.24$\pm$0.35&	91.44$\pm$0.19&	93.00$\pm$0.08&	72.28$\pm$0.56&	85.57$\pm$0.28\\	\cline{2-11}
										
& \bf \textsc{D-Trilinear} &	90.43$\pm$0.23&	73.57$\pm$0.17&	69.50$\pm$0.32&	84.15$\pm$0.23&	79.41$\pm$0.24&	91.54$\pm$0.13&	93.19$\pm$0.07&	72.76$\pm$0.18&	85.83$\pm$0.13\\
										
& \bf \textsc{D-Quadlinear} &\textbf{90.44$\pm$0.07}&	\textbf{75.05$\pm$0.35}&\textbf{70.49$\pm$0.68}&	\textbf{84.41$\pm$0.29}&\textbf{80.09$\pm$0.35}&	91.97$\pm$0.14&93.35$\pm$0.05&	\textbf{74.42$\pm$0.24}&\textbf{86.58$\pm$0.14}\\

& \bf \textsc{D-Pentalinear} &	90.29$\pm$0.06&	74.09$\pm$0.49&	69.90$\pm$0.45&	83.81$\pm$0.12&	79.52$\pm$0.28&	\textbf{91.98$\pm$0.09}&	\textbf{93.43$\pm$0.07}&	74.17$\pm$0.29&	86.53$\pm$0.15\\
\hline \hline
\multirow{8}{*}{\rotatebox{90}{\bf \textsc{Word \& Char}}}
& \bf \textsc{BiLSTM-LAN}&90.71$\pm$0.20&77.18$\pm$0.28&77.83$\pm$0.90&83.97$\pm$0.84&82.42$\pm$0.55&91.84$\pm$0.10&94.20$\pm$0.09&72.33$\pm$0.16&86.12$\pm$0.12\\	\cline{2-11}
& \bf \textsc{Softmax} &	90.39$\pm$0.11&	76.87$\pm$0.26&	77.40$\pm$0.32&	83.89$\pm$0.36&	82.14$\pm$0.26&	91.13$\pm$0.08&	94.02$\pm$0.05&	71.32$\pm$0.08&	85.49$\pm$0.07\\
										
& \bf \textsc{Vanilla CRF} &	91.15$\pm$0.22&	78.13$\pm$0.36&	79.65$\pm$1.52&	85.45$\pm$0.55&	83.59$\pm$0.66&	91.59$\pm$0.12&	94.23$\pm$0.05&	73.34$\pm$0.07&	86.39$\pm$0.08\\	\cline{2-11}
										
& \bf \textsc{TwoBilinear} &	90.98$\pm$0.10&	77.84$\pm$0.41&	79.12$\pm$0.85&	85.48$\pm$0.30&	83.36$\pm$0.42&	91.78$\pm$0.11&	93.99$\pm$0.07&	72.07$\pm$0.26&	85.94$\pm$0.15\\
										
& \bf \textsc{ThreeBilinear} &	91.24$\pm$0.16&	77.48$\pm$0.53&	80.15$\pm$0.33&	85.27$\pm$0.10&	83.53$\pm$0.28&	91.75$\pm$0.06&	93.92$\pm$0.13&	72.20$\pm$0.44&	85.95$\pm$0.21\\
										
& \bf \textsc{Trilinear} &	91.30$\pm$0.11&	77.41$\pm$0.25&	79.69$\pm$0.81&	85.60$\pm$0.32&	83.50$\pm$0.38&	91.70$\pm$0.24&	94.14$\pm$0.11&	72.42$\pm$0.59&	86.08$\pm$0.31\\	\cline{2-11}
										
& \bf \textsc{D-Trilinear} &	91.18$\pm$0.18&	77.98$\pm$0.45&	80.02$\pm$0.68&	\textbf{85.83$\pm$0.25}&	83.75$\pm$0.39&	91.97$\pm$0.17&	94.24$\pm$0.10&	73.05$\pm$0.15&	86.42$\pm$0.14\\
										
& \bf \textsc{D-Quadlinear} &	\textbf{91.34$\pm$0.12}&	\textbf{78.89$\pm$0.29}&	80.81$\pm$0.78&	85.75$\pm$0.39&	\textbf{84.20$\pm$0.39}&\textbf{92.36$\pm$0.06}&94.52$\pm$0.02&	74.33$\pm$0.23&87.07$\pm$0.10\\

& \bf \textsc{D-Pentalinear} &	91.08$\pm$0.32&	78.53$\pm$0.57&	\textbf{80.99$\pm$0.55}&	85.42$\pm$0.23&	84.01$\pm$0.42&	92.28$\pm$0.14&	\textbf{94.58$\pm$0.04}&\textbf{74.48$\pm$0.41}&\textbf{87.11$\pm$0.20}\\

										
										
\end{tabular}
\caption{\label{table:details of NER & Chunking Result}
Results on NER and Chunking tasks. BiLSTM-LAN \cite{cui-zhang-2019-hierarchically} is one of the current state-of-the-art sequence labeling approaches.}
\end{table*}

\begin{table*}
\setlength\tabcolsep{3.8pt}
\renewcommand\arraystretch{1.36}
\scriptsize
\centering
\begin{tabular}{l|l|ccccccccc}
 \multicolumn{2}{c|}{}& \textbf{Chinese} & \textbf{Dutch} &  \textbf{English}& \textbf{German} & \textbf{Hindi} &  \textbf{Indonesian}&\textbf{Italian} &  \textbf{Japanese}& \bf Avg.\\
\hline \hline
\multirow{8}{*}{\rotatebox{90}{\bf \textsc{Word Embedding}}}
& \bf \textsc{BiLSTM-LAN} &93.34$\pm$0.09&94.44$\pm$0.14&\textbf{95.09$\pm$0.08}&\textbf{94.18$\pm$0.12}&\textbf{97.02$\pm$0.10}&\textbf{91.76$\pm$0.35}&\textbf{97.62$\pm$0.20}&94.54$\pm$0.28&94.75$\pm$0.10\\	\cline{2-11}
& \bf \textsc{Softmax} &93.13$\pm$0.05&94.11$\pm$0.04&	94.66$\pm$0.09&	93.43$\pm$0.09&	96.72$\pm$0.06&	91.00$\pm$0.06&	97.34$\pm$0.07&	95.39$\pm$0.26&94.47$\pm$0.09\\
& \bf \textsc{Vanilla CRF} &93.10$\pm$0.08&94.14$\pm$0.14&	94.70$\pm$0.04&	93.41$\pm$0.13&	96.74$\pm$0.05&	91.23$\pm$0.05&	97.41$\pm$0.05&95.50$\pm$0.24&94.53$\pm$0.10\\	\cline{2-11}
										
& \bf \textsc{TwoBilinear} &93.05$\pm$0.05&94.16$\pm$0.09&	94.64$\pm$0.08&	93.40$\pm$0.13&	96.71$\pm$0.05&	91.11$\pm$0.07&	97.38$\pm$0.05&95.41$\pm$0.10&94.48$\pm$0.08\\
& \bf \textsc{ThreeBilinear} &93.06$\pm$0.07&94.21$\pm$0.11&	94.66$\pm$0.05&	93.43$\pm$0.11&	96.72$\pm$0.05&	91.07$\pm$0.09&	97.33$\pm$0.10&95.41$\pm$0.16&94.49$\pm$0.09\\
& \bf \textsc{Trilinear}&93.64$\pm$0.15&94.35$\pm$0.07&94.61$\pm$0.08&	93.38$\pm$0.12&	96.89$\pm$0.06&	90.99$\pm$0.13&	97.58$\pm$0.05&95.44$\pm$0.21&94.61$\pm$0.11\\	\cline{2-11}
										
& \bf \textsc{D-Trilinear} &93.71$\pm$0.12&94.35$\pm$0.07&	94.77$\pm$0.10&	93.52$\pm$0.20&	96.93$\pm$0.04&	91.27$\pm$0.11&97.59$\pm$0.04&95.46$\pm$0.18&94.70$\pm$0.11
\\
& \bf \textsc{D-Quadlinear} &\textbf{94.21$\pm$0.11}&94.59$\pm$0.08&94.85$\pm$0.09&93.84$\pm$0.13&\textbf{97.02$\pm$0.06}&91.36$\pm$0.06&\textbf{97.62$\pm$0.08}&	\textbf{95.78$\pm$0.17}&\textbf{94.91$\pm$0.10}\\
& \bf \textsc{D-Pentalinear} &	93.91$\pm$0.21&	\textbf{94.60$\pm$0.06}&	94.78$\pm$0.12&	93.54$\pm$0.18&	97.01$\pm$0.13&	91.46$\pm$0.12&	97.51$\pm$0.13&	95.60$\pm$0.23&	94.80$\pm$0.15\\

\hline \hline
\multirow{8}{*}{\rotatebox{90}{\bf \textsc{Word \& Char}}}	
& \bf \textsc{BiLSTM-LAN}&94.00$\pm$0.12&95.45$\pm$0.12&95.86$\pm$0.15&\textbf{94.72$\pm$0.06}&97.10$\pm$0.02&\textbf{93.80$\pm$0.03}&\textbf{98.14$\pm$0.09}&96.34$\pm$0.06&95.68$\pm$0.08\\\cline{2-11}
& \bf \textsc{Softmax} &93.67$\pm$0.07&	95.28$\pm$0.12&	95.92$\pm$0.02&	94.28$\pm$0.11&	96.96$\pm$0.10&	93.48$\pm$0.06&	97.88$\pm$0.04&97.14$\pm$0.14&95.58$\pm$0.08\\
& \bf \textsc{Vanilla CRF} &93.55$\pm$0.16&	95.37$\pm$0.14&	96.01$\pm$0.04&	94.28$\pm$0.13&	96.96$\pm$0.05&	93.51$\pm$0.05&	97.94$\pm$0.05&97.21$\pm$0.17&95.60$\pm$0.10\\	\cline{2-11}
										
& \bf \textsc{TwoBilinear} &93.51$\pm$0.06&	95.23$\pm$0.10&	95.96$\pm$0.07&	94.43$\pm$0.17&	96.94$\pm$0.08&	93.48$\pm$0.19&	97.88$\pm$0.11&97.20$\pm$0.07&95.58$\pm$0.11\\
& \bf \textsc{ThreeBilinear} &93.50$\pm$0.10&	95.25$\pm$0.10&	95.92$\pm$0.09&	94.47$\pm$0.10&	96.89$\pm$0.04&	93.35$\pm$0.15&	97.84$\pm$0.04&97.08$\pm$0.11&95.54$\pm$0.09\\
& \bf \textsc{Trilinear} &93.93$\pm$0.14&	95.25$\pm$0.06&	95.84$\pm$0.09&	94.43$\pm$0.11&	97.02$\pm$0.11&	93.33$\pm$0.13&	98.01$\pm$0.06&97.22$\pm$0.26&95.63$\pm$0.12\\	\cline{2-11}
										
& \bf \textsc{D-Trilinear} &94.03$\pm$0.12&	95.30$\pm$0.10&	96.04$\pm$0.05&	94.34$\pm$0.06&	97.06$\pm$0.02&	93.45$\pm$0.12&98.09$\pm$0.03&97.21$\pm$0.17&95.69$\pm$0.08
\\
& \bf \textsc{D-Quadlinear} &\textbf{94.58$\pm$0.10}&	\textbf{95.50$\pm$0.16}&96.01$\pm$0.13&94.51$\pm$0.09&97.19$\pm$0.06&93.53$\pm$0.13&98.02$\pm$0.04&\textbf{97.25$\pm$0.08}&\textbf{95.82$\pm$0.10}\\
& \bf \textsc{D-Pentalinear} &	94.50$\pm$0.21&	95.45$\pm$0.13&	\textbf{96.08$\pm$0.12}&	94.29$\pm$0.18&	\textbf{97.20$\pm$0.12}&	93.60$\pm$0.12&	97.98$\pm$0.05&	97.23$\pm$0.08&	95.79$\pm$0.13\\
\hline \hline
\multirow{7}{*}{\rotatebox{90}{\bf \textsc{BERT Embedding}}}
& \bf \textsc{Softmax} &96.70$\pm$0.13&	96.19$\pm$0.06&	96.67$\pm$0.04&	95.06$\pm$0.12&	96.75$\pm$0.03&	92.90$\pm$0.12&	98.34$\pm$0.05&	96.84$\pm$0.13&	96.18$\pm$0.08\\
& \bf \textsc{Vanilla CRF} &96.72$\pm$0.10&	96.20$\pm$0.12&	96.80$\pm$0.08&	95.30$\pm$0.03&	96.74$\pm$0.08&	92.90$\pm$0.08&	98.31$\pm$0.16&	96.85$\pm$0.10&	96.23$\pm$0.09\\	\cline{2-11}
										
& \bf \textsc{TwoBilinear} &	96.54$\pm$0.09&	96.02$\pm$0.19&	96.78$\pm$0.09&	95.33$\pm$0.05&	96.71$\pm$0.04&	92.88$\pm$0.10&	98.36$\pm$0.06&	96.86$\pm$0.08&	96.18$\pm$0.09\\
& \bf \textsc{ThreeBilinear} &	96.58$\pm$0.10&	95.96$\pm$0.06&	\textbf{96.85$\pm$0.06}&	95.13$\pm$0.07&	96.68$\pm$0.06&	92.81$\pm$0.14&	98.24$\pm$0.06&	96.84$\pm$0.05&	96.14$\pm$0.08\\		& \bf \textsc{Trilinear} &	96.71$\pm$0.03&	96.01$\pm$0.29&	96.84$\pm$0.42&	95.26$\pm$0.09&	96.68$\pm$0.03&	92.70$\pm$0.10&	98.25$\pm$0.11&	96.75$\pm$0.06&	96.17$\pm$0.14\\	\cline{2-11}
										
& \bf \textsc{D-Trilinear} &	96.79$\pm$0.06&	96.24$\pm$0.18&	96.82$\pm$0.10&	95.33$\pm$0.03&	96.79$\pm$0.06&	92.88$\pm$0.05&	98.33$\pm$0.08&	96.85$\pm$0.07&	96.25$\pm$0.08\\
& \bf \textsc{D-Quadlinear} &	\textbf{96.86$\pm$0.06}&	\textbf{96.25$\pm$0.09}&	\textbf{96.85$\pm$0.06}&	95.30$\pm$0.04&	\textbf{96.87$\pm$0.03}&	\textbf{92.98$\pm$0.10}&	\textbf{98.37$\pm$0.07}&	\textbf{97.05$\pm$0.11}&	\textbf{96.32$\pm$0.07}\\
& \bf \textsc{D-Pentalinear} &	96.84$\pm$0.03&	96.20$\pm$0.13&	96.81$\pm$0.18&	\textbf{95.50$\pm$0.11}&	96.85$\pm$0.04&	92.93$\pm$0.16&	98.35$\pm$0.09&	96.98$\pm$0.13&	96.31$\pm$0.11\\
\end{tabular}
\caption{\label{table:details of Coarse-grained POS Result}
Results on Coarse POS task. BiLSTM-LAN \cite{cui-zhang-2019-hierarchically} is one of the current state-of-the-art sequence labeling approaches.
}
\end{table*}

\begin{table*}
\setlength\tabcolsep{3.8pt}
\renewcommand\arraystretch{1.36}
\scriptsize
\centering
\begin{tabular}{l|l|ccccccccc}
 \multicolumn{2}{c|}{}& \textbf{Chinese} & \textbf{Dutch} &  \textbf{English}& \textbf{German} & \textbf{Hindi} &  \textbf{Indonesian}&\textbf{Italian} &  \textbf{Japanese}& \bf Avg.\\ 
\hline \hline
\multirow{8}{*}{\rotatebox{90}{\bf \textsc{Word Embedding}}}
& \bf \textsc{BiLSTM-LAN} &93.16$\pm$0.07&90.45$\pm$0.25&\textbf{94.60$\pm$0.12}&\textbf{96.41$\pm$0.04}&\textbf{96.54$\pm$0.02}&94.11$\pm$0.11&\textbf{97.62$\pm$0.05}&92.69$\pm$0.45&94.45$\pm$0.14\\	\cline{2-11}
& \bf \textsc{Softmax} &93.33$\pm$0.09&92.17$\pm$0.11&	94.40$\pm$0.05&	96.22$\pm$0.03&	96.25$\pm$0.08&	94.66$\pm$0.03&	97.26$\pm$0.10&	95.02$\pm$0.14&94.91$\pm$0.08\\	
& \bf \textsc{Vanilla CRF}&93.22$\pm$0.11&92.13$\pm$0.09&	94.41$\pm$0.09&96.22$\pm$0.06&	96.39$\pm$0.09&	94.60$\pm$0.05&	97.27$\pm$0.06&94.86$\pm$0.12&94.89$\pm$0.08\\	\cline{2-11}
										
& \bf \textsc{TwoBilinear} &93.19$\pm$0.13&92.03$\pm$0.09&	94.17$\pm$0.05&	96.15$\pm$0.04&	96.34$\pm$0.08&	94.72$\pm$0.06&	97.24$\pm$0.04&	94.67$\pm$0.35&94.81$\pm$0.11\\	
& \bf \textsc{ThreeBilinear} &93.29$\pm$0.09&92.12$\pm$0.06&	94.12$\pm$0.10&	96.20$\pm$0.05&	96.39$\pm$0.06&	94.82$\pm$0.12&	97.30$\pm$0.06&94.73$\pm$0.14&94.87$\pm$0.09\\	
& \bf \textsc{Trilinear} &93.79$\pm$0.08&91.91$\pm$0.17&	94.20$\pm$0.09&	96.27$\pm$0.05&	96.42$\pm$0.10&	94.60$\pm$0.11&	97.49$\pm$0.12&	94.80$\pm$0.32&94.94$\pm$0.13\\	\cline{2-11}
										
& \bf \textsc{D-Trilinear} &93.78$\pm$0.02&92.21$\pm$0.28&	94.29$\pm$0.09&	96.27$\pm$0.08&	96.46$\pm$0.05&	94.77$\pm$0.09&	97.50$\pm$0.05&	\textbf{95.25$\pm$0.10}&95.07$\pm$0.10\\	
& \bf \textsc{D-Quadlinear} &\textbf{94.24$\pm$0.07}&\textbf{92.36$\pm$0.18}&94.36$\pm$0.06&	96.28$\pm$0.07&	\textbf{96.54$\pm$0.07}&\textbf{94.94$\pm$0.09}&	\textbf{97.62$\pm$0.03}&95.20$\pm$0.06&\textbf{95.19$\pm$0.08}\\	
& \bf \textsc{D-Pentalinear} &94.12$\pm$0.18&	92.21$\pm$0.22&	94.30$\pm$0.34&	96.25$\pm$0.46&	96.39$\pm$0.08&	94.70$\pm$0.09&	97.55$\pm$0.10&	95.03$\pm$0.08&	95.07$\pm$0.19\\
\hline \hline
\multirow{8}{*}{\rotatebox{90}{\bf \textsc{Word \& Char}}}				
& \bf \textsc{BiLSTM-LAN}&93.88$\pm$0.11&92.38$\pm$0.45&95.58$\pm$0.08&\textbf{97.14$\pm$0.05}&96.69$\pm$0.03&94.80$\pm$0.05&\textbf{98.02$\pm$0.01}&94.75$\pm$0.23&95.41$\pm$0.13\\	\cline{2-11}
& \bf \textsc{Softmax} &93.86$\pm$0.10&93.05$\pm$0.10&	95.59$\pm$0.08&	97.07$\pm$0.06&	96.55$\pm$0.03&	94.93$\pm$0.06&	97.85$\pm$0.05&96.85$\pm$0.07&95.72$\pm$0.07\\	
& \bf \textsc{Vanilla CRF}&93.69$\pm$0.08&\textbf{93.21$\pm$0.14}&	95.58$\pm$0.11&	97.08$\pm$0.11&	96.59$\pm$0.03&	94.80$\pm$0.08&	97.89$\pm$0.06&96.75$\pm$0.27&95.70$\pm$0.11\\	\cline{2-11}

& \bf \textsc{TwoBilinear} &93.54$\pm$0.08&92.91$\pm$0.22&	95.60$\pm$0.13&	97.03$\pm$0.05&	96.59$\pm$0.06&	94.93$\pm$0.05&	97.83$\pm$0.09&	96.67$\pm$0.11&95.64$\pm$0.10\\	
& \bf \textsc{ThreeBilinear}&93.62$\pm$0.08&93.03$\pm$0.13&	95.55$\pm$0.05&	97.09$\pm$0.06&	96.61$\pm$0.04&	94.97$\pm$0.09&	97.79$\pm$0.06&96.63$\pm$0.19&95.66$\pm$0.09\\	
& \bf \textsc{Trilinear}&94.02$\pm$0.07&92.99$\pm$0.29&	95.58$\pm$0.09&	97.10$\pm$0.03&	96.61$\pm$0.07&	94.84$\pm$0.11&	97.98$\pm$0.06&	96.59$\pm$0.15&95.71$\pm$0.11\\	\cline{2-11}

& \bf \textsc{D-Trilinear}&94.05$\pm$0.15&92.83$\pm$0.17&	\textbf{95.69$\pm$0.04}&97.11$\pm$0.05&96.64$\pm$0.02&	94.95$\pm$0.08&	97.96$\pm$0.03&	96.80$\pm$0.08&95.75$\pm$0.08\\	
& \bf \textsc{D-Quadlinear}&\textbf{94.49$\pm$ 0.11}&93.03$\pm$0.18&	95.64$\pm$0.05&	97.10$\pm$0.03&\textbf{96.78$\pm$0.05}&95.04$\pm$0.09&\textbf{98.02$\pm$0.05}&96.90$\pm$0.07&\textbf{95.88$\pm$0.08}\\
& \bf \textsc{D-Pentalinear} &	94.20$\pm$0.12&	93.03$\pm$0.20&	95.63$\pm$0.02&	97.04$\pm$0.05&	96.70$\pm$0.02&	\textbf{95.08$\pm$0.05}&	97.91$\pm$0.30&	\textbf{96.93$\pm$0.12}&	95.82$\pm$0.11\\

\hline \hline
\multirow{7}{*}{\rotatebox{90}{\bf \textsc{BERT Embedding}}}								
& \bf \textsc{Softmax} &	96.54$\pm$0.05&	93.58$\pm$0.13&	96.39$\pm$0.05&	97.52$\pm$0.06&	96.26$\pm$0.10&	91.56$\pm$0.06&	98.24$\pm$0.03&	96.16$\pm$0.06&	95.83$\pm$0.07\\	
& \bf \textsc{Vanilla CRF} &	96.53$\pm$0.11&	93.57$\pm$0.12&	96.38$\pm$0.08&	97.47$\pm$0.03&	96.31$\pm$0.08&	\textbf{91.75$\pm$0.10}&	98.24$\pm$0.09&	96.25$\pm$0.11&	95.81$\pm$0.09\\	\cline{2-11}
& \bf \textsc{TwoBilinear} &	96.47$\pm$0.05&	93.45$\pm$0.15&	\textbf{96.42$\pm$0.05}&	97.52$\pm$0.06&	96.32$\pm$0.06&	\textbf{91.75$\pm$0.15}&	98.16$\pm$0.04&	96.24$\pm$0.15&	95.79$\pm$0.09\\	
& \bf \textsc{ThreeBilinear} &	96.45$\pm$0.10&	93.26$\pm$0.21&	96.32$\pm$0.07&	97.50$\pm$0.06&	96.27$\pm$0.03&	91.73$\pm$0.20&	98.18$\pm$0.04&	96.19$\pm$0.18&	95.74$\pm$0.11\\	
& \bf \textsc{Trilinear} &	96.60$\pm$0.08&	93.53$\pm$0.13&	96.22$\pm$0.05&	97.55$\pm$0.06&	96.22$\pm$0.08&	90.97$\pm$0.24&	98.22$\pm$0.07&	96.09$\pm$0.14&	95.67$\pm$0.11\\	\cline{2-11}
& \bf \textsc{D-Trilinear} &	96.68$\pm$0.07&	93.68$\pm$0.19&	96.31$\pm$0.08&	\textbf{97.57$\pm$0.06}&	96.37$\pm$0.07&	91.55$\pm$0.09&	98.26$\pm$0.04&	96.35$\pm$0.09&	95.74$\pm$0.11\\	
& \bf \textsc{D-Quadlinear} &	96.74$\pm$0.10&	\textbf{93.69$\pm$0.21}&	96.39$\pm$0.05&	97.56$\pm$0.03&	\textbf{96.41$\pm$0.07}&	91.65$\pm$0.18&	\textbf{98.29$\pm$0.02}&	\textbf{96.44$\pm$0.05}&	\textbf{95.90$\pm$0.09}\\
& \bf \textsc{D-Pentalinear} &	\textbf{97.10$\pm$0.02}&	93.40$\pm$0.03&	96.40$\pm$0.01&	97.54$\pm$0.09&	96.33$\pm$0.11&	91.46$\pm$0.20&	\textbf{98.29$\pm$0.06}&	96.30$\pm$0.08&	95.85$\pm$0.08
\\
\end{tabular}
\caption{\label{table:details of Fine-grained POS Result}
Results on Fine POS task. BiLSTM-LAN \cite{cui-zhang-2019-hierarchically} is one of the current state-of-the-art sequence labeling approaches.
}
\end{table*}

\end{document}